\documentclass[conference]{IEEEtran}
\IEEEoverridecommandlockouts
\usepackage{cite}
\usepackage{amsmath,amssymb,amsfonts}
\usepackage{algorithmic}
\usepackage{graphicx}
\usepackage{textcomp}
\usepackage{xcolor}
\usepackage{lipsum}
\usepackage{multirow}
\def\BibTeX{{\rm B\kern-.05em{\sc i\kern-.025em b}\kern-.08em
    T\kern-.1667em\lower.7ex\hbox{E}\kern-.125emX}}
\begin{document}

\title{{Pixel Wised Lesion Prediction on COVID-19 CT Imagery: A Comparative Analysis of Automated Image Segmentation Architectures}\\
}

\author{
\IEEEauthorblockN{Sarmad Khan}
\IEEEauthorblockA{\textit{Dept. of Computer \& Software Engg.}\\
\textit{Nat. Univ. of Sciences \& Technology}\\
Islamabad, Pakistan\\
rdxesss@gmail.com}
\and
\IEEEauthorblockN{Basim Azam}
\IEEEauthorblockA{\textit{School of Computing \& Info. Systems}\\
\textit{University of Melbourne}\\
Melbourne, Australia\\
basim.azam@unimelb.edu.au}
\and
\IEEEauthorblockN{Arslan Shaukat}
\IEEEauthorblockA{\textit{Dept. of Computer \& Software Engg.}\\
\textit{Nat. Univ. of Sciences \& Technology}\\
Islamabad, Pakistan\\
arslanshaukat@ceme.nust.edu.pk}
}

\maketitle

\begin{abstract}
In recent years, there has been a notable increase in the level of attention that is given to algorithms based on deep learning in the context of medical image segmentation.  Nevertheless, the reliability of the field has been hindered due to the absence of a standardized methodology for performance analysis and the utilization of different datasets in previous research. The primary objective of the research is to comprehensively evaluate contemporary segmentation frameworks combined with state-of-the-art pre-trained backbones in order to accurately predict COVID-19 lesions in CT images. Moreover, this evaluation can serve as a point of reference for the segmentation of images in various other imaging scenarios. In order to accomplish this, we integrate four distinct deep learning architectures, namely Unet, PSPNet, Linknet, and FPN, with six pre-trained encoders, including VGG 19, DenseNet 121, Inception ResNet V2, MobileNet V2, SeresNet 101, and EfficientNet B0. This approach enables the development of diverse testing architectures. In the context of image segmentation, our research encompassed both binary and multi-class experimentation. The findings derived from our analysis of three distinct COVID-19 CT segmentation datasets indicate that deep learning architectures yield precise and efficient segmentation outcomes. Significantly, a maximum F1-Score of 98\% was attained for binary class segmentation, while multi-class segmentation yielded F1-Scores of 75\% and 77\% across two separate datasets. The utilization of artificial intelligence and deep learning enhances the diagnostic process for pandemic diseases across multiple dimensions.
\end{abstract}

\begin{IEEEkeywords}
COVID-19, Deep learning, Image Segmentation, CT, X-rays
\end{IEEEkeywords}

\section{Introduction}
The COVID-19 virus initially emerged in December 2019, and its global recognition as a significant pandemic occurred by March 2020 \cite{covid19}. Due to the extremely transmissible nature of the disease,
conventional measures such as social distancing have demonstrated limited efficacy, thus highlighting the imperative for prompt and accurate detection methods. Computed tomography scans, a widely utilized imaging modality known for its high-resolution capabilities, provide comprehensive pathological data with a particular emphasis on the lungs. CT scan images of individuals diagnosed with COVID-19 often exhibit ground glass opacity within the affected region, a characteristic that is not discernible in conventional X-ray scans \cite{wang2020clinical}. The Reverse-Transcription Polymerase Chain Reaction (RT-PCR) method has been employed for the diagnosis of COVID-19. Nevertheless, studies have indicated that this technique exhibits lower reliability compared to CT scans and entails a longer turnaround time for obtaining the test results \cite{ai2020correlation}.
\par
Healthcare professionals often encounter the challenge of reviewing a substantial volume of computed tomography (CT) scans, a process that is protracted and can potentially lead to errors. To address this issue, researchers are currently developing automatic segmentation algorithms that utilize high-resolution CT scans to accurately delineate regions of interest (ROIs) with diverse sizes and shapes. Healthcare professionals employ targeted strategies to enhance their diagnostic process by directing their attention toward specific regions of interest rather than considering the entire clinical picture. The prior research reveals a significant dearth of research investigations that have conducted statistical analysis on deep learning architectures aimed at both binary and multi-class segmentation of COVID-19 images employing CT scans.
. The primary intent of the research is to provide a thorough experimental analysis that compares the performance of various deep learning architectures on binary and multi-class segmentation of COVID-19 CT benchmark datasets.
\par
The following sections are organized in the subsequent sequence: Section II provides a comprehensive overview of pertinent field research. Section III provides an in-depth analysis and evaluation of the proposed architecture. In Section IV, the experiments and their corresponding outcomes are examined and analyzed. Section V of the research encompasses an analysis of the findings, while Section VI provides the concluding section of the research.

\section{Related Work}
A multitude of artificial intelligence methodologies have been employed for the segmentation of the coronavirus. The purpose of this literature review is to provide an in-depth investigation of the utilization of machine learning and deep learning techniques in the field of COVID-19 image segmentation, with particular emphasis on addressing the difficulties posed by the coronavirus. Additionally, it will provide a review of the aforementioned subjects and methodologies. The experiments were conducted utilizing publicly accessible datasets, which consisted of CT scan images.

\par
% Narges Saeedizadeh et al. proposed a unique method for segmenting coronaviruses on CT scans, using U-Net as the foundational architecture \cite{saeedizadeh2021covid}. With a 99 percent mIOU rate and an 86 percent Dice score, the model was able to reliably identify COVID-19-related lung regions in the experiment. Two deep learning networks, "U-Net" and "SegNet," \cite{saood2021covid} were proposed as a framework. The results of the experiments demonstrate that SegNet surpasses the other approaches by 95 percent in accuracy when classifying infected and non-infected regions.

Saeedizadeh et al introduced a novel approach to segment coronaviruses on CT scans, employing U-Net as the underlying architecture \cite{saeedizadeh2021covid}. The model demonstrated a high level of accuracy in identifying COVID-19-related lung regions during the experiment, as evidenced by a mIOU rate of 99 percent and a Dice score of 86 percent. The framework proposed by Saood et al. (2021) introduced two deep learning networks, namely "U-Net" and "SegNet" \cite{saood2021covid}. The experimental findings indicate that SegNet exhibits superior performance compared to alternative methodologies, achieving a 95 percent increase in accuracy for the classification of infected and non-infected regions.

\par
% A baseline architecture of a traditional 3D U-Net was introduced by D. Müller et al \cite{muller2020automated}. The suggested method outperforms current COVID-19 segmentation approaches and enhances medical image analysis with minimal data, with Dice similarity values of 95\% for lungs and 76\% for infection. Athanasios \cite{voulodimos2021deep} investigated how well DL architectures worked for COVID-19 segmentation. Even though the dataset has an uneven number of classes and mistakes that were made by humans when annotating the data, the result shows that fully convolutional neural networks (FCNN) can correctly segment, with an F1-score of 65\%.

D. Müller et al. (2020) introduced a foundational structure for a conventional 3D U-Net in their work \cite{muller2020automated}. The proposed methodology demonstrates superior performance compared to existing COVID-19 segmentation techniques, thereby improving the analysis of medical images with limited data. The Dice similarity coefficients achieved are 95\% for lung segmentation and 76\% for infection segmentation. The study conducted by Voulodimos et al \cite{voulodimos2021deep} examined the efficacy of deep learning architectures in the context of COVID-19 segmentation. Despite the presence of an imbalanced class distribution and human annotation errors within the dataset, the findings indicate that fully convolutional neural networks (FCNN) are capable of accurately segmenting, achieving an F1-score of 65\%.

\par
% An architecture that is termed "dual branch combination network" (often abbreviated as DCN) \cite{gao2021dual} was designed and constructed for coronavirus diagnosis using a large-scale dataset of 1,202 participants from 10 Chinese research institutions, accomplishing independent-level classification and segmentation at the same time. On both the internal and external datasets, the recommended DCN model was more accurate than other models, with 96 and 92 percent accuracy, respectively. MiniSeg \cite{qiu2020miniseg} was proposed as a lightweight deep learning model for image segmentation, with an IOU rate of 85 percent.

The authors of the paper developed and implemented a network architecture known as the "dual branch combination network" (DCN) \cite{gao2021dual} for the purpose of diagnosing coronavirus. This architecture was designed to handle a substantial dataset consisting of 1,202 participants from 10 research institutions in China. The DCN demonstrated the ability to perform both classification and segmentation tasks simultaneously, achieving results comparable to those obtained through independent-level analysis. The DCN model demonstrated superior accuracy compared to other models on both the internal and external datasets, achieving accuracy rates of 96\% and 92\% respectively. The MiniSeg model, as proposed by Qiu et al. (2020) \cite{qiu2020miniseg} is a deep learning architecture designed for image segmentation tasks. It is specifically designed to be lightweight, making it suitable for resource-constrained environments. The model achieves an Intersection over Union (IOU) rate of 85 percent, indicating its effectiveness in accurately delineating object boundaries in segmented images.

\par
% Improved Dilation with a Dense Network of Attention Gates Convolution-UNET, developed by Alex Novel \cite{raj2021adid}, is a new COVID-19 pulmonary infection segmentation system. The research indicates the proposed ADID-UNET model can be able to consistently segment COVID-19 lung-infected regions with accuracy, specificity, and dice coefficient efficiency metrics that are all more than 80\%. In addition, the suggested method fared better than previous state-of-the-art designs, scoring 80\% and 82\%, respectively, on the DC and F1 scales. This was the case when the methods were compared.

Enhanced Dilation via a Dense Network of Attention Gates The Convolution-UNET, a novel system developed by Alex Novel \cite{raj2021adid}, represents a recent advancement in the field of COVID-19 pulmonary infection segmentation. According to the findings of the study, the ADID-UNET model demonstrates consistent performance in segmenting regions of lung infection caused by COVID-19. The model achieves high levels of accuracy, specificity, and dice coefficient efficiency, all-surpassing 80\%. Furthermore, the proposed approach exhibited superior performance compared to previous cutting-edge models, achieving accuracy scores of 80\% and 82\% on the DC and F1 metrics, respectively. This was the situation observed when the methods were compared.

\par
% While most deep learning architectures have shown superiority over their predecessors in COVID-19 image segmentation, comprehensive experimentation and comparability with other state-of-the-art approaches are still needed. To the best of our knowledge, no prior study has compared COVID-19 binary and multi-class segmentation analyses using various datasets. As a result, this research gives a full experimental evaluation of COVID-19 lesion segmentation on a CT benchmark dataset, as well as a comparison of pre-trained feature extraction networks with a number of modern segmentation heads.

Although numerous deep learning architectures have demonstrated their superiority in COVID-19 image segmentation compared to previous methods, there is still a need for extensive experimentation and comparison with other state-of-the-art approaches to ensure a comprehensive evaluation. To our knowledge, there is no existing research that has conducted a comparative analysis of COVID-19 binary and multi-class segmentation using diverse datasets. Consequently, this study provides a comprehensive experimental assessment of COVID-19 lesion segmentation using a CT benchmark dataset. Additionally, it includes a comparative analysis of pre-trained feature extraction networks in conjunction with various contemporary segmentation heads.

\section{Proposed Architecture}
This section provides an overview of the architectures and encoders utilized for addressing segmentation problems, along with the datasets and data augmenting methods used.

\subsection{Datasets}
The COVID-19 image segmentation problem made use of the following publicly available CT scan datasets associated with COVID-19. Figures 1 and 2 depict visual representations of computed tomography scans and corresponding masks acquired from their respective datasets.

\begin{itemize}
    \item The initial dataset \cite{medicalseg} comprises a set of 100 axial CT scans. The scans utilized in this study were obtained from openly accessible JPG images, which were specifically sourced from a cohort of over 40 individuals who had been diagnosed with COVID-19. The radiologist employed a three-label segmentation approach to classify the images, assigning the labels of ground glass, consolidation, and lungs. 
    \item The next dataset \cite{medicalseg} comprises complete volumes containing both negative and positive slices. Among the total of 29 slices, 373 have been evaluated as positive and subsequently segmented by the radiologist. The term "ground glass opacity" (GGO) pertains to the presence of indistinct, grayish areas that can be observed in computed tomography (CT) scans or X-rays of the lungs. The areas of increased density observed in the lungs are depicted by the grey patches. Lung consolidation occurs when the alveoli, which are the tiny air sacs in the lungs, become filled with substances other than air. These substances can include fluids like pus, blood, or water, as well as solids such as gastric contents or cellular matter.
    \item The dataset referenced in \cite{zenodo} consists of a total of 20 computed tomography (CT) images of patients diagnosed with COVID-19. Additionally, the dataset includes expertly segmented images of the lungs and areas affected by infections. The radiologist utilized three distinct labels to perform image segmentation: "Joint Lung and Infection Mask" and "Independent Lungs + Infection Mask."
\end{itemize}

\begin{figure}
\centerline{\includegraphics[width=\textwidth,width=9cm,]{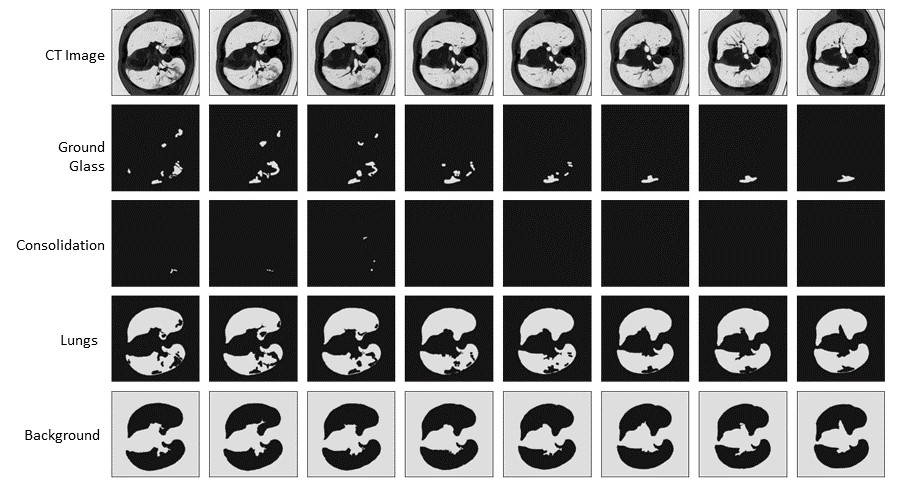}}
\caption{CT images and masks of Medical Segmentation Dataset}
\label{fig:1}
\end{figure}

\begin{figure}
\centerline{\includegraphics[width=\textwidth,width=9cm,]{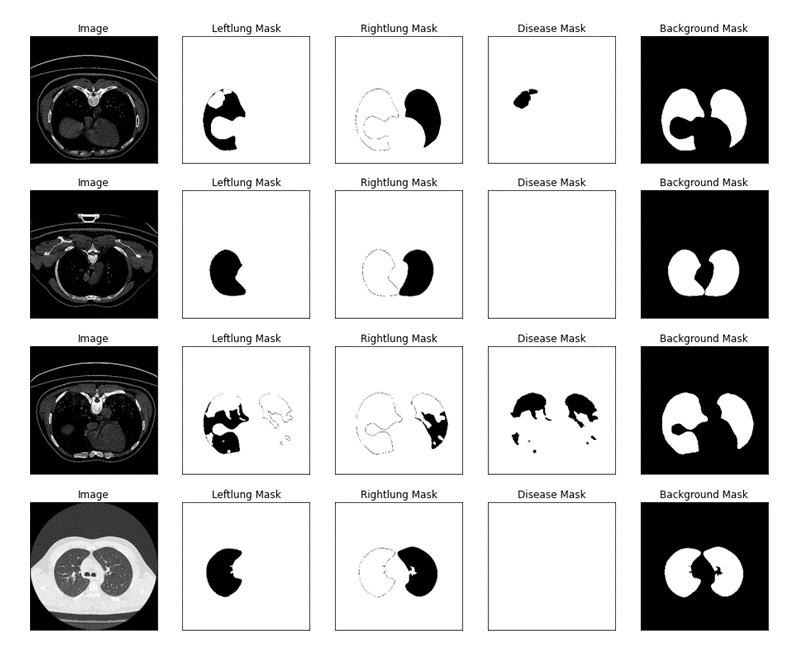}}
\caption{CT images and masks of Zenodo Dataset}
\label{fig:2}
\end{figure}

\subsection{Data Augmentation}
In order to address the limited availability of COVID-CT segmented images and masks, the "albumentation library" was employed for the purposes of binary and multi-class segmentation. This involved the utilization of a synthetic data creation approach within the framework of data augmentation. Several image processing techniques were utilized, including blurring, sharpening, horizontal flips, rotation, Gaussian noise, cropping, shift range, and scaling. Table I presents information regarding the number of images within the dataset, as well as the allocation of these images for training and testing purposes.

\begin{table}[htbp]
\centering
\caption{ Dataset Train Test Split Information}
\begin{tabular}{|c|c|c|c|} 
\hline
\textbf{Dataset}                                                                             & \textbf{Number of Images}                           & \begin{tabular}[c]{@{}c@{}}\textbf{Training}\\\textbf{Images}\end{tabular} & \textbf{Testing Images}                             \\ 
\hline
Zenodo                                                                                       & 20                                                  & 3520                                                                       & 500                                                 \\ 
\hline
\begin{tabular}[c]{@{}c@{}}Medical \\Segmentation\\(Radiopedia\\+\\Medseg part)\end{tabular} & \begin{tabular}[c]{@{}c@{}}829\\+\\100\end{tabular} & \begin{tabular}[c]{@{}c@{}}704\\+\\85\end{tabular}                         & \begin{tabular}[c]{@{}c@{}}125\\+\\15\end{tabular}  \\
\hline
\end{tabular}
\end{table}

\subsection{Proposed Architecture}
% The main architectures used for image segmentation are: \textit{"Unet, PSPNet (Pyramid Scene Parsing Network), Linknet, FPN (Feature Pyramid Network)"}
The primary architectural deep learning frameworks employed for image segmentation include \textit{"Unet, PSPNet (Pyramid Scene Parsing Network), Linknet, FPN (Feature Pyramid Network)"}

\par

\begin{figure}
\centerline{\includegraphics[width=\textwidth,width=9cm,]{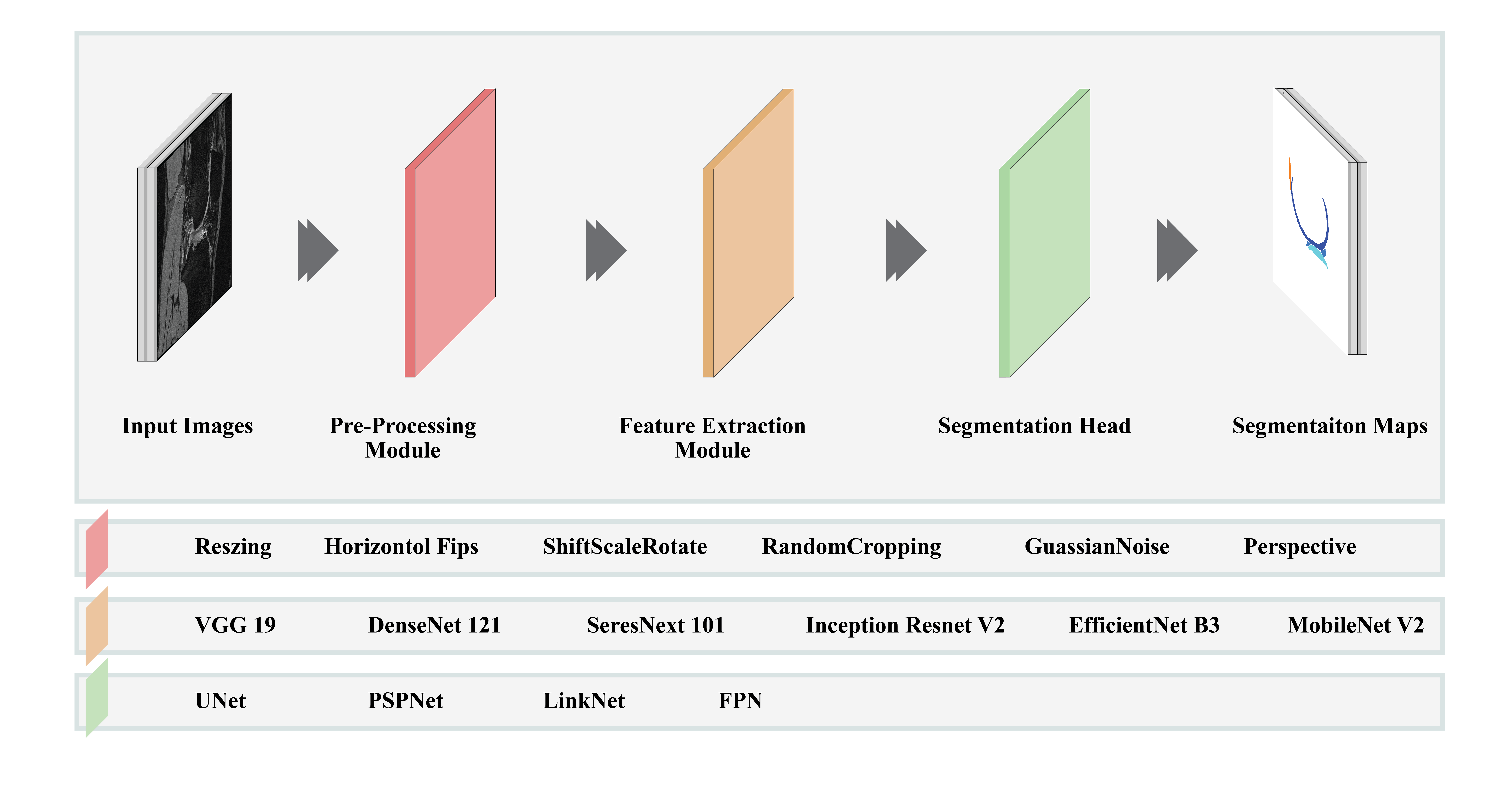}}
\caption{Proposed Architecture for COVID-19 Segmentation}
\label{fig:3}
\end{figure}

The encoder of Unet, as proposed by Ronneberger et al. \cite{ronneberger2015u}, is designed to reduce the dimensionality of features in order to accommodate a smaller bottleneck. Conversely, the decoder of Unet is responsible for restoring the original proportions of these features. Additionally, skip connections have been employed in order to enhance the process of image segmentation, as demonstrated in previous studies. The PSPNet \cite{zhao2017pyramid} employs a pyramid pooling module to effectively combine global context information from the image, accompanied by a corresponding loss function. The encoder and decoder networks employed in Linknet \cite{chaurasia2017linknet} share the same architecture as those in Unet. However, a notable distinction lies in the utilization of residual blocks instead of convolutional blocks within the encoder and decoder networks. The Feature Pyramid Network (FPN) proposed by Lin et al \cite{lin2017feature} is a similar concept to Unet, but it differs in its approach to incorporating and merging features. Instead of directly copying and attaching features, FPN utilizes a 1x1 convolution layer where the features are added.

The segmentation architectures also utilize pre-trained feature extraction networks as encoders, which include \textit{"Efficient Net, MobileNet V2, Seresnet 101, Densenet 121, VGG-19, and Inception Resnet V2"}.

\par
The Visual Geometry Group (VGG) \cite{simonyan2014very}, which was the proposer and achieved the second position in the 2014 ImageNet Competition, is commonly referred to as VGG \cite{deng2009imagenet}. The aforementioned approach was among the initial ones to emphasize the significance of depth in deep learning. Due to its straightforward repetitive framework, it is commonly selected for tasks such as feature extraction. The MobileNet architecture \cite{howard2017mobilenets} was specifically designed to address the computational limitations of devices with slight computing capabilities, such as embedded devices and mobile phones. In contrast, the DenseNet architecture \cite{huang2017densely} utilizes links between each layer and every other layer in a feed-forward manner. The compound scaling module was proposed as a viable method for uniformly scaling the depth, breadth, and resolution in EfficientNet \cite{tan2019efficientnet}. EfficientNet is an enhanced version of MobileNet, while SeresNet \cite{hu2018squeeze} is a variant of ResNet that incorporates squeeze-and-excitation blocks to facilitate dynamic channel-wise feature recalibration within the network. The utilization of Inception networks \cite{szegedy2015going} allows for efficient computation through the reduction of dimensionality. The primary objective of the modules is to effectively mitigate issues related to computational expenses and overfitting.

Table II provides the specifications of the hyperparameters employed in the deep learning network utilized for COVID-19 image segmentation. Additionally, Figure 3 illustrates the proposed architecture for COVID-19 image segmentation.

\begin{table}
\centering
\caption{Model Hyperparameters for Segmentation}
\begin{tabular}{|c|c|c|c|} 
\hline
\textbf{Framework}                                                  & \begin{tabular}[c]{@{}c@{}}\textbf{Depth of }\\\textbf{Encoder}\end{tabular} & \begin{tabular}[c]{@{}c@{}}\textbf{Batch }\\\textbf{Normalization.}\end{tabular} & \textbf{Variations}                                                                                                \\ 
\hline
Unet                                                                & 5                                                                            & \begin{tabular}[c]{@{}c@{}}Yes\\(decoder)\end{tabular}                  & \begin{tabular}[c]{@{}c@{}}decoder channel \\size =\\ (256,128,64,32,16)\end{tabular}                              \\ 
\hline
Link  Net                                                           & 5                                                                            & \begin{tabular}[c]{@{}c@{}}Yes\\(decoder)\end{tabular}                  & —                                                                                                                  \\ 
\hline
\begin{tabular}[c]{@{}c@{}}Pyramid Scene\\Parsing Net\end{tabular}  & 3                                                                            & \begin{tabular}[c]{@{}c@{}}Yes\\(decoder)\end{tabular}                  & \begin{tabular}[c]{@{}c@{}}dropout=0.2\\output channels=512\end{tabular}                                           \\ 
\hline
\begin{tabular}[c]{@{}c@{}}Feature \\Pyramid~\\Network\end{tabular} & 5                                                                            & No                                                                      & \begin{tabular}[c]{@{}c@{}}dropout=0.2\\merge policy=add\\segment channels=128\\pyramid channels=256\end{tabular}  \\
\hline
\end{tabular}
\end{table}

\section{Experiments \& Results}

\subsection{Model Hyperparameters}
Each segmentation model was formed using a distinct combination of four architectures and six encoders. The Sigmoid activation function was employed for binary segmentation, whereas SoftMax was employed for multi-class segmentation. The Keras library, specifically version 2.3.1, and the TensorFlow library, specifically version 2.1.0, were utilized. The experiment employed a batch size of 16 and utilized the ADAM optimizer with a learning rate of 0.0001 to optimize the loss function. The approach also involved the utilization of the combo loss, which is a combination of weighted binary cross-entropy, focal loss, and Tversky loss. Extensive image augmentation was conducted prior to the commencement of training. The binary class segmentation utilized an image resolution of 256 x 256, while the multi-class segmentation employed image resolutions of 256 x 256 (Zenodo) and 512 x 512 (Medical Segmentation). The Python script was executed utilizing Google's Research online Jupyter Notebook-based service, Collaboratory, and the utilization of the Tesla K80 GPU, which is accessible via Google Collaboratory, was also employed to augment the speed of performance.

\subsection{Evaluation Metrics}
The model’s performance was evaluated using the 
Jaccard Index \& F1-score.
\par
\textbf{Jaccard Index}, also known as the Intersection over Union (IoU), is a quantitative measure employed to assess the level of overlap between the predicted and the actual masks for segmentation in the context of image segmentation tasks.

\begin{equation}
Jaccard Index = \frac{TP}{TP+FN+FP}
\label{eq:jaccard}
\end{equation}

\textbf{F1-Score} is a widely used metric in the field of image segmentation to evaluate the precision and recall of an algorithm in accurately recognizing and demarcating specific areas or objects within an image.

\begin{equation}
F1-Score = \frac{2TP}{2TP+FP+FN}
\label{eq:f1}
\end{equation}

\subsection{Results}
The experiments for image segmentation were carried out by conducting three experiments using two distinct datasets.
 The first experiment focused on binary segmentation and utilized the Zenodo dataset. The remaining two experiments involved multi-class segmentation and employed both the Zenodo dataset and the medical segmentation dataset. The results are succinctly presented in Table IV.

\begin{figure}
\centerline{\includegraphics[width=\textwidth,width=9cm,]{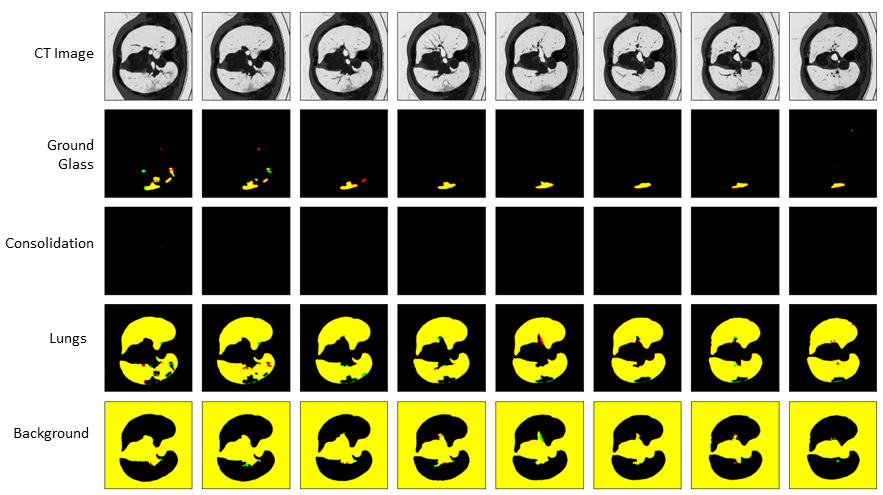}}
\caption{COVID-19 Segmented Images in terms of Pixel Wise Accuracy for Medical Segmentation Dataset}
\label{fig:4}
\end{figure}

\begin{figure}
\centerline{\includegraphics[width=\textwidth,width=9cm,]{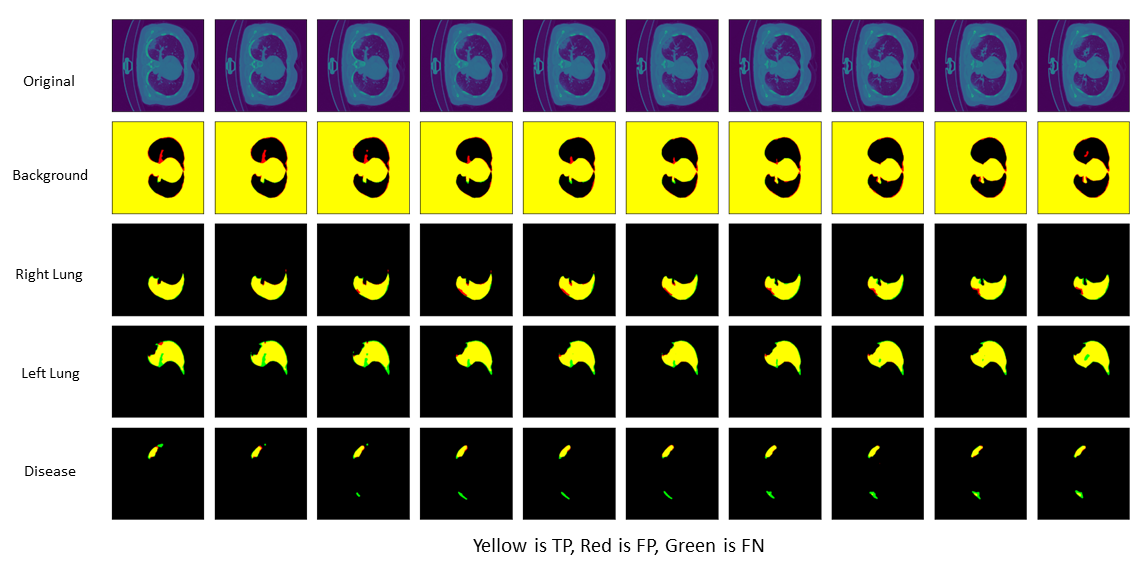}}
\caption{COVID-19 Segmented Images in terms of Pixel Wise Accuracy for Zenodo Dataset}
\label{fig:5}
\end{figure}

\begin{figure}
\centerline{\includegraphics[width=\textwidth,width=9.0cm,height=5cm]{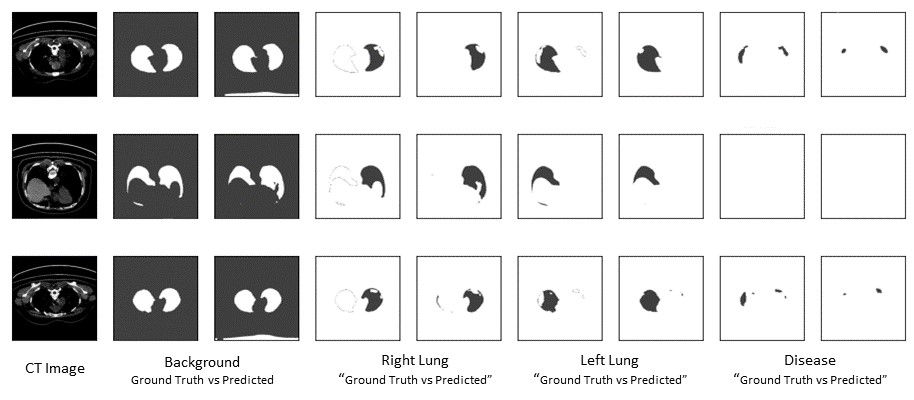}}
\caption{Visual Appearance of COVID-19 Ground Truth vs Predicted CT Images}
\label{fig:6}
\end{figure}

The Zenodo dataset was utilized to perform binary class segmentation, wherein two distinct classes were identified and segmented, namely "background" and "disease." Upon careful analysis of Table IV, it becomes evident that all architectures utilizing respected backbones demonstrated the highest Jaccard Index and F1-Score. However, it is worth noting that the Unet architecture outperformed the others, achieving an impressive overall F1-Score of 98\% on the test dataset. 

\par
In the same way, two experiments were undertaken for the purpose of multi-class segmentation. These experiments were conducted on distinct datasets, each containing unique images and corresponding masks. The initial experiment conducted on the medical segmentation dataset involved the segmentation of four distinct classes, namely "ground glass," "consolidation," "lungs," and "background." Table IV provides an in-depth assessment, revealing that architectures utilizing various backbones exhibited superior F1-Scores for multi-class segmentation. Notably, the Linknet architecture, employing Inception Resnet V2 as the encoder, demonstrated exceptional performance, achieving an impressive F1-Score of 75\% on the test dataset. In the subsequent experiment conducted on the Zenodo dataset, the segmentation process involved the segmentation of four distinct classes, namely "background," "right lung," "left lung," and "disease." Upon closer analysis of Table IV, it becomes evident that architectures utilizing respected backbones demonstrated superior performance in terms of Jaccard Index scores and F1-Scores. Notably, the FPN architecture with Densenet 121 as the encoder exhibited exceptional proficiency in segmenting masks for image segmentation on the test dataset. This particular architecture achieved an impressive F1-Score of 77\% on the test dataset.
\par
Figures 4 and 5 present a visual comparison in terms of pixel-wise accuracy, while Figure 6 illustrates a comparison between the ground truth images and predicted images of COVID-19 CT scans.

\begin{table}[htbp]
\centering
\caption{Comparison of the proposed model with various state-of-the-art baseline segmentation approaches}
\begin{tabular}{|c|c|c|c|c|} 
\hline
\textbf{ Mode }                                                              & \textbf{ Dataset }                                                              & \textbf{ Author } & \textbf{ Architecture }                                                                             & \begin{tabular}[c]{@{}c@{}}\textbf{Results }\\\textbf{F1-Score}\end{tabular}  \\ 
\hline
\multirow{5}{*}{\begin{tabular}[c]{@{}c@{}}Binary Class \\Seg.\end{tabular}} & \multirow{3}{*}{Zenodo}                                                         & {[}7]             & FCN                                                                                                 & 0.65                                                                              \\ 
\cline{3-5}
                                                                             &                                                                                 & {[}8]             & DCN                                                                                                 & 0.92                                                                              \\ 
\cline{3-5}
                                                                             &                                                                                 & \textbf{Proposed} & \begin{tabular}[c]{@{}c@{}}\textbf{Unet. }\\\textbf{Mobilenet }\\\textbf{V2}\end{tabular}           & \textbf{0.98}                                                                     \\ 
\cline{2-5}
                                                                             & \multirow{2}{*}{\begin{tabular}[c]{@{}c@{}}Medical \\Segmentation\end{tabular}} & {[}4]             & TV-Unet                                                                                             & 0.86                                                                              \\ 
\cline{3-5}
                                                                             &                                                                                 & {[}5]             & SegNet                                                                                              & 0.82                                                                              \\ 
\hline
\multirow{6}{*}{\begin{tabular}[c]{@{}c@{}}Multi-class \\Seg.\end{tabular}}  & \multirow{3}{*}{Zenodo}                                                         & {[}6]             & 3D-Unet                                                                                             & 0.76                                                                              \\ 
\cline{3-5}
                                                                             &                                                                                 & {[}9]             & MiniSeg                                                                                             & 0.76                                                                              \\ 
\cline{3-5}
                                                                             &                                                                                 & \textbf{Proposed} & \begin{tabular}[c]{@{}c@{}}\textbf{FPN. }\\\textbf{Densenet 121}\end{tabular}                       & \textbf{0.77}                                                                     \\ 
\cline{2-5}
                                                                             & \multirow{3}{*}{\begin{tabular}[c]{@{}c@{}}Medical \\Segmentation\end{tabular}} & {[}5]             & Unet                                                                                                & 0.55                                                                              \\ 
\cline{3-5}
                                                                             &                                                                                 & {[}9]             &  Miniseg                                                                                                   & 0.75                                                                              \\ 
\cline{3-5}
                                                                             &                                                                                 & \textbf{Proposed} & \begin{tabular}[c]{@{}c@{}}\textbf{Linknet. }\\\textbf{Inception}\\\textbf{~Resnet V2}\end{tabular} & \textbf{0.75}                                                                     \\
\hline
\end{tabular}
\end{table}

%%%%%%%%%%%%%%%%%%%%%%%%%%%%%%%%%%%%%%%%%%%%%%%%%%%%%%%%%%%%%%%%%%%%%%%%%%%%%%%%%%%%%%%%%%%%%%%%%%%%%%%%%%%%%%%%%%%%%%%%%%%%%%%%%%%%%%%%%%%%%%%%%%%%%%%%%%%%%%%%%%%%%%%%%%%%%%%%%%%%%%%%%%%%%%%%%%%%%%%%%%%%%%%%%%%%%%%%%%%%%%%%%%%%%%%%%%%%%%%%%%%%%%%%%%%%%

\begin{table*}[!htbp]
% \fontsize{8.5}{10.5}\selectfont
\tiny
\setlength{\tabcolsep}{4pt}
\renewcommand{\arraystretch}{1.5}  % Increased spacing between rows
\centering
\caption{Evaluation Metrics}
\begin{tabular}{|p{2.4cm}|p{2.4cm}|p{2.4cm}|p{2.4cm}|p{2.4cm}|p{2.4cm}|}
\hline
\textbf{Dataset} & \textbf{Class} & \textbf{Architecture} & \textbf{Backbone} & \textbf{Jaccard Index} & \textbf{F1-Score} \\ 
\hline
\multirow{24}{*}{Zenodo}               & \multirow{24}{*}{Binary class} & \multirow{6}{*}{Unet}    & Efficient Net B0              & 0.943                  & 0.970              \\ 
\cline{4-6}
                                       &                                &                          & \textbf{Mobile Net V2}       & \textbf{0.972}         & \textbf{0.986}     \\ 
\cline{4-6}
                                       &                                &                          & Inception Resnet V2          & 0.970                  & 0.985              \\ 
\cline{4-6}
                                       &                                &                          & Densenet 121                & 0.943                  & 0.970              \\ 
\cline{4-6}
                                       &                                &                          & Seresnet 101                & 0.927                  & 0.962              \\ 
\cline{4-6}
                                       &                                &                          & VGG 19                      & 0.915                  & 0.955              \\ 
\cline{3-6}
                                       &                                & \multirow{6}{*}{PSPNet}  & Efficientnet B0             & 0.927                  & 0.962              \\ 
\cline{4-6}
                                       &                                &                          & Mobilenet V2                & 0.932                  & 0.964              \\ 
\cline{4-6}
                                       &                                &                          & Inception Resnet V2          & 0.956                  & 0.977              \\ 
\cline{4-6}
                                       &                                &                          & Densenet 121                & 0.957                  & 0.977              \\ 
\cline{4-6}
                                       &                                &                          & Seresnet 101                & 0.915                  & 0.955              \\ 
\cline{4-6}
                                       &                                &                          & VGG 19                      & 0.854                  & 0.921              \\ 
\cline{3-6}
                                       &                                & \multirow{6}{*}{LinkNet} & Efficient Net B0              & 0.921                  & 0.958              \\ 
\cline{4-6}
                                       &                                &                          & Mobilenet V2                & 0.961                  & 0.980              \\ 
\cline{4-6}
                                       &                                &                          & Inception Resnet V2          & 0.961                  & 0.980              \\ 
\cline{4-6}
                                       &                                &                          & Densenet 121                & 0.915                  & 0.955              \\ 
\cline{4-6}
                                       &                                &                          & Seresnet 101                & 0.949                  & 0.973              \\ 
\cline{4-6}
                                       &                                &                          & VGG 19                      & 0.957                  & 0.977              \\ 
\cline{3-6}
                                       &                                & \multirow{6}{*}{FPN}     & Efficient Net B0              & 0.921                  & 0.958              \\ 
\cline{4-6}
                                       &                                &                          & Mobilenet V2                & 0.940                  & 0.969              \\ 
\cline{4-6}
                                       &                                &                          & Inception Resnet V2          & 0.959                  & 0.979              \\ 
\cline{4-6}
                                       &                                &                          & Densenet 121                & 0.928                  & 0.962              \\ 
\cline{4-6}
                                       &                                &                          & Seresnet 101                & 0.965                  & 0.982              \\ 
\cline{4-6}
                                       &                                &                          & VGG 19                      & 0.886                  & 0.939              \\ 
\hline
\multirow{24}{*}{Zenodo}               & \multirow{24}{*}{Multi class~} & \multirow{6}{*}{Unet}    & Efficient Net B0              & 0.66                   & 0.72               \\ 
\cline{4-6}
                                       &                                &                          & Mobilenet V2                & 0.65                   & 0.72               \\ 
\cline{4-6}
                                       &                                &                          & Inception Resnet V2          & 0.67                   & 0.73               \\ 
\cline{4-6}
                                       &                                &                          & Densenet 121                & 0.68                   & 0.74               \\ 
\cline{4-6}
                                       &                                &                          & Seresnet 101                & 0.69                   & 0.75               \\ 
\cline{4-6}
                                       &                                &                          & VGG 19                      & 0.64                   & 0.70               \\ 
\cline{3-6}
                                       &                                & \multirow{6}{*}{PSPNet}  & Efficient Net B0              & 0.53                   & 0.60               \\ 
\cline{4-6}
                                       &                                &                          & Mobilenet V2                & 0.48                   & 0.54               \\ 
\cline{4-6}
                                       &                                &                          & Inception Resnet V2          & 0.64                   & 0.71               \\ 
\cline{4-6}
                                       &                                &                          & Densenet 121                & 0.66                   & 0.72               \\ 
\cline{4-6}
                                       &                                &                          & Seresnet 101                & 0.59                   & 0.67               \\ 
\cline{4-6}
                                       &                                &                          & VGG 19                      & 0.67                   & 0.74               \\ 
\cline{3-6}
                                       &                                & \multirow{6}{*}{Linknet} & Efficient Net B0              & 0.68                   & 0.74               \\ 
\cline{4-6}
                                       &                                &                          & Mobilenet V2                & 0.65                   & 0.71               \\ 
\cline{4-6}
                                       &                                &                          & Inception Resnet V2          & 0.69                   & 0.76               \\ 
\cline{4-6}
                                       &                                &                          & Densenet 121                & 0.56                   & 0.63               \\ 
\cline{4-6}
                                       &                                &                          & Seresnet 101                & 0.63                   & 0.69               \\ 
\cline{4-6}
                                       &                                &                          & VGG 19                      & 0.58                   & 0.64               \\ 
\cline{3-6}
                                       &                                & \multirow{6}{*}{FPN}     & Efficient Net B0              & 0.64                   & 0.70               \\ 
\cline{4-6}
                                       &                                &                          & Mobilenet V2                & 0.62                   & 0.69               \\ 
\cline{4-6}
                                       &                                &                          & Inception Resnet V2          & 0.69                   & 0.76               \\ 
\cline{4-6}
                                       &                                &                          & \textbf{Densenet 121}       & \textbf{0.70}          & \textbf{0.77}      \\ 
\cline{4-6}
                                       &                                &                          & Seresnet 101                & 0.68                   & 0.74               \\ 
\cline{4-6}
                                       &                                &                          & VGG 19                      & 0.68                   & 0.74               \\ 
\cline{1-1}\cline{3-6}
\multirow{24}{*}{Medical Segmentation} &                                & \multirow{6}{*}{Unet}    & Efficient Net B0              & 0.64                   & 0.72               \\ 
\cline{4-6}
                                       &                                &                          & Mobilenet V2                & 0.63                   & 0.69               \\ 
\cline{4-6}
                                       &                                &                          & Inception Resnet V2          & 0.66                   & 0.74               \\ 
\cline{4-6}
                                       &                                &                          & Densenet 121                & 0.60                   & 0.68               \\ 
\cline{4-6}
                                       &                                &                          & Seresnet 101                & 0.65                   & 0.73               \\ 
\cline{4-6}
                                       &                                &                          & VGG 19                      & 0.60                   & 0.67               \\ 
\cline{3-6}
                                       &                                & \multirow{6}{*}{PSPNet}  & Efficient Net B0              & 0.62                   & 0.69               \\ 
\cline{4-6}
                                       &                                &                          & Mobilenet V2                & 0.53                   & 0.61               \\ 
\cline{4-6}
                                       &                                &                          & Inception Resnet V2          & 0.64                   & 0.72               \\ 
\cline{4-6}
                                       &                                &                          & Densenet 121                & 0.66                   & 0.73               \\ 
\cline{4-6}
                                       &                                &                          & Seresnet 101                & 0.66                   & 0.74               \\ 
\cline{4-6}
                                       &                                &                          & VGG 19                      & 0.56                   & 0.64               \\ 
\cline{3-6}
                                       &                                & \multirow{6}{*}{Linknet} & Efficient Net B0              & 0.66                   & 0.74               \\ 
\cline{4-6}
                                       &                                &                          & Mobilenet V2                & 0.65                   & 0.74               \\ 
\cline{4-6}
                                       &                                &                          & \textbf{Inception Resnet V2} & \textbf{0.68}          & \textbf{0.75}      \\ 
\cline{4-6}
                                       &                                &                          & Densenet 121                & 0.60                   & 0.68               \\ 
\cline{4-6}
                                       &                                &                          & Seresnet 101                & 0.66                   & 0.73               \\ 
\cline{4-6}
                                       &                                &                          & VGG 19                      & 0.61                   & 0.68               \\ 
\cline{3-6}
                                       &                                & \multirow{6}{*}{FPN}     & Efficient Net B0              & 0.65                   & 0.72               \\ 
\cline{4-6}
                                       &                                &                          & Mobilenet V2                & 0.60                   & 0.67               \\ 
\cline{4-6}
                                       &                                &                          & Inception Resnet V2          & 0.65                   & 0.72               \\ 
\cline{4-6}
                                       &                                &                          & Densenet 121                & 0.66                   & 0.74               \\ 
\cline{4-6}
                                       &                                &                          & Seresnet 101                & 0.65                   & 0.73               \\ 
\cline{4-6}
                                       &                                &                          & VGG 19                      & 0.58                   & 0.65               \\
\hline
\end{tabular}
\end{table*}

%%%%%%%%%%%%%%%%%%%%%%%%%%%%%%%%%%%%%%%%%%%%%%%%%%%%%%%%%%%%%%%%%%%%%%%%%%%%%%%%%%%%%%%%%%

%%%%%%%%%%%%%%%%%%%%%%%%%%%%%%%%%%%%%%%%%%%%%%%%%%%%%%%%%%%%%%%%%%%%%%%%%%%%%%%%%%%%%%%%%%%

\section{Discussion}
The motivation behind our research stems from the limited availability of the datasets, comprehensive experiments, and baseline comparisons among the numerous distinct frameworks currently in existence. The results of this research can serve as a compendium of cutting-edge benchmark models for binary and multi-class segmentation of COVID-19. It can be employed by deep learning model developers and researchers to assess the performance of their novel models by comparing metrics against the same training/validation dataset and/or parameters. The primary objective of this study was to provide a comprehensive assessment of deep learning architectures utilized in computer-aided design (CAD) systems, specifically those capable of employing image segmentation techniques to accurately delineate the region of interest in CT images associated with COVID-19. The utilization of deep learning architectures has been noted to potentially offer advantages to radiologists in accurately and expeditiously identifying cases of COVID-19 through the analysis of radiological imaging within a reduced time frame. The process of segmentation was conducted utilizing four distinct deep-learning architectures in conjunction with six encoders. The findings indicate that deep learning architectures demonstrate consistent segmentation of COVID-19 images. The Unet architecture, utilizing MobileNet V2 as its backbone, showed strong performance in binary class segmentation. On the other hand, Linknet-Inception Resnet V2 and FPN-Densenet 121 showed notable performance in multi-class segmentation. Table III presents a comparison between our proposed models and other state-of-the-art segmentation approaches as documented in the existing literature.

\section{Conclusion}
Given the current global impact of COVID-19, which has culminated in significant loss of life, the objective of this research was to evaluate and compare various deep learning architectures in the context of medical image segmentation. In order to mitigate the limited availability of minimalistic training data and fulfill the efficiency demands of computer-aided design implementation, we conducted a thorough evaluation and conducted an extensive and meticulous comparison. The findings indicate that the Unet architecture demonstrated strong performance in binary class segmentation, achieving an F1 score of 98\%. Additionally, the Linknet and FPN architectures achieved F1 scores of 75\% and 77\% respectively in multi-class segmentation. Deep learning architectures have the potential to assist radiologists in accurately and expeditiously identifying COVID-19 through the utilization of radiological imaging, thereby reducing the time required for computer-aided detection systems.

%%%%%%%%%%%%%%%%%%%%%%%%%%%%%%%%%%%%%%%%%%%%%%%%%%%%%%%%%%%%%%%%%%%%%%%%%%%%%%%%%%%%%%%%%%%%%%%%%%%%%%%%%%%%%%%%%%%%%%%%%%%%%%%%%%%%%%%%%%%%%%%%%%%%%%%%%%%%%%%%%%%%%%%%%%%%%%%%%%%%%%%%%%%%%%%%%%%%%%%%%%%%%%%%%%%%%%%%%%%%%%%%%%%%%%%%%%%%%%%%%%%%%%%%%%%%%%%%%%%%%%%%%%%%%%%%%%%%%%%%%%%%%%%%%%%%%%%%%%%%%%%%%%%%%%%%%%%%%%%%%%%%%%%%%%%%%%%%%%%%%%%%%%%%%%%%%%%%%%%%%%%%%%%%%%%%%%%%%%%%%%%%%%%%%%%%%%%%%%%%%%%%%%%%%%%%%%%%%%%%%%%%%%%%%%%%%%%%%%%%%%%%%%%%%%%%%%%%%%%%%%%%%%%%%%%%%%%%%%%%%%%%%%%%%%%%%%%%%%%%%%%%%%%%%%%%%%%%%%%%%%%%%%%%%%%%%%%%%%%%%%%%%%%%%%%%%%%%%%%%%%%%%%%%%%%%%%%%%%%%%%%%%%%%%%%%%%%%%%%%%%%%%%%%%%%%%%%%%%%%%%%%%%%%%%%%%%%%%%%%%%%%%%%%%%%%%%%%%%%%%%%%%%%%%%%%%%%%%%%%%%%%%%%%%%%%%%%%%%%%%%%%%%%%%%%%%%%%%%%%%%%%%%%%%%%%%%%%%%%%%%%%%%%%%%%%%%%%%%%%%%%%%%%%%%%%%%%%%%%%%%%%%%%%%%%%%%%%%%%%%%%%%%%%%%%%%%%%%%%%%%%%%%%%%%%%%%%%%%%%%%%%%%%%%%%%%%%%%%%%%%%%%%%%%%%%%%%%%%%%%%%%%%%%%

\end{document}